%% file: main.tex
\pdfoutput=1

\documentclass[11pt]{article}

\usepackage[final]{acl}

\usepackage{times}
\usepackage{latexsym}

\usepackage[T1]{fontenc}

\usepackage[utf8]{inputenc}

\usepackage{microtype}

\usepackage{inconsolata}

\usepackage{graphicx}
\usepackage{amsmath}
\usepackage{cleveref}
\usepackage{xspace}
\usepackage{multirow}
\usepackage{booktabs}
\usepackage{xcolor}

\definecolor{paircolor}{HTML}{4682B4}

\newcommand\myfont[1]{\smash{{\usefont{T1}{qag}{m}{n}#1}}}
\newcommand{\modelnamefancy}{\myfont{ACON}\xspace}
\newcommand{\modelname}{ACON\xspace}
\newcommand{\modelnamelong}{Any-to-any CONsistency\xspace}

\title{Are Any-to-Any Models More Consistent\\Across Modality Transfers Than Specialists?}

\author{
 \textbf{Jiwan Chung} \quad
 \textbf{Janghan Yoon} \quad
 \textbf{Junhyeong Park} \quad
 \textbf{Sangeyl Lee} \\
 \textbf{Joowon Yang} \quad
 \textbf{Sooyeon Park} \quad
 \textbf{Youngjae Yu}
 \\
 Yonsei University
\\
\\
\texttt{jiwan.chung.research@gmail.com}
}

\begin{document}
\maketitle
\begin{abstract}
\input{sections/abstract}
\end{abstract}

\input{sections/introduction}

\input{sections/method}

\input{sections/data}

\input{sections/implementation}

\input{sections/experiments}

\input{sections/related_work}

\input{sections/conclusion}

\input{sections/ack}

\bibliography{custom}

\clearpage
\appendix

\input{sections/ax_setups}

\input{sections/ax_data}

\input{sections/ax_exp}

\input{sections/ax_prompts}

\end{document}

%% file: sections/abstract.tex
Any-to-any generative models aim to enable seamless interpretation and generation across multiple modalities within a unified framework, yet their ability to preserve relationships across modalities remains uncertain. Do unified models truly achieve cross-modal coherence, or is this coherence merely perceived? To explore this, we introduce \textsc{\modelnamefancy}, a dataset of 1,000 images (500 newly contributed) paired with captions, editing instructions, and Q\&A pairs to evaluate cross-modal transfers rigorously. Using three consistency criteria—cyclic consistency, forward equivariance, and conjugated equivariance—our experiments reveal that any-to-any models do not consistently demonstrate greater cross-modal consistency than specialized models in pointwise evaluations such as cyclic consistency. However, equivariance evaluations uncover weak but observable consistency through structured analyses of the intermediate latent space enabled by multiple editing operations. We release our code and data at \url{https://github.com/JiwanChung/ACON}.

%% file: sections/introduction.tex
\section{Introduction}
\label{sec:intro}

Any-to-any generative models are designed to both interpret and generate multiple modalities—such as text, images, and audio—within a unified framework~\cite{wang2022ofa,lu2024unified,wang2024emu3}. In contrast to modality-specific approaches which often rely on textual interfaces to mediate generation~\cite{openai2023gpt4,erwold-2024-qwen2vl-flux}, any-to-any models share the majority of parameters across different modalities. This design choice suggests potential advantages in flexibility and transferability between modalities.

However, the practical value of any-to-any models remains uncertain. At their current stage of development, they often fail to consistently outperform specialized models~\cite{podellsdxl,blackforest2024flux,liu2024improved} in terms of output quality. Also, they may be less training-efficient due to the significant computational overhead of optimizing a single, large-scale system. As a result, it remains unclear whether such models confer tangible benefits over their modality-specific counterparts.

\input{figs/teaser}

What, then, should we anticipate from any-to-any models? Prior work~\cite{huang2021makes,lu2023theory} has proposed viewing multimodal learning as an attempt to approximate a shared latent representation from each modality’s partial view. Building on this perspective, we posit that if a single any-to-any model successfully learns such a unified latent space, it should produce more coherent cross-modal conversions than two separate modality-specific models, each constrained by its own distinct latent approximation.

\input{figs/fig_main}

To verify this conjecture, we test whether any-to-any models achieve greater consistency in cross-modal transfer than pairs of modality-specific models. We formalize this consistency using three criteria: \textit{cyclic consistency}, requiring that converting an input from text to image and back again recovers the original input; \textit{forward equivariance}, ensuring that applying a modification before or after cross-modal conversion yields the same result; and \textit{conjugated equivariance}, providing an alternative formulation of equivariance that utilizes both text-to-image and image-to-text conversions.

We introduce \modelnamefancy (\modelnamelong), a meticulously annotated dataset to assess coherence in cross-modal transformations. 
It comprises 1,000 images, including 500 private images specifically contributed for this study. Each image is paired with a human-written dense caption aimed at faithful reconstruction, three image-editing instructions, and ten binary question-answer pairs for evaluating output similarity. In addition, every editing instruction is accompanied by two prompt-conditioned Q\&As to capture the effects of the transformation.

Experiments on \modelname reveal that any-to-any models do not consistently exhibit greater cross-modal consistency compared to arbitrary combinations of specialized models, particularly in pointwise evaluations such as cyclic consistency. However, equivariance evaluations demonstrate that weak consistency between text-to-image and image-to-text capabilities can be observed in distributional analyses of the intermediate latent space, enabled by multiple editing operations.

We anticipate that \modelname will serve as a diagnostic benchmark to evaluate the benefits and trade-offs of training any-to-any models within a unified framework: any-to-any models do not show great consistency at the current stage.
We will release our code and data to support further research and development in this field.

%% file: figs/teaser.tex
\begin{figure}
    \centering
    \includegraphics[width=0.48\textwidth]{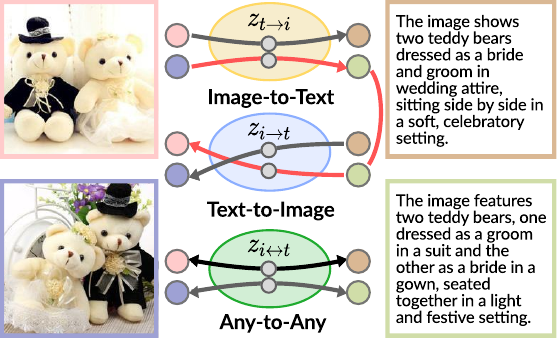}
    \caption{
    We examine the consistency of any-to-any models compared to separate image-to-text and text-to-image models. An effective any-to-any model, capable of learning a unified latent space $z$, is expected to mitigate issues like cyclic consistency failures, as depicted by the red lines. The illustration is a conceptual case drawn with images from MMVP~\cite{tong2024eyes}.
    }
    \label{fig:intuition}
\end{figure}

%% file: figs/fig_main.tex
\begin{figure*}[!ht]
    \centering
    \includegraphics[width=0.98\textwidth]{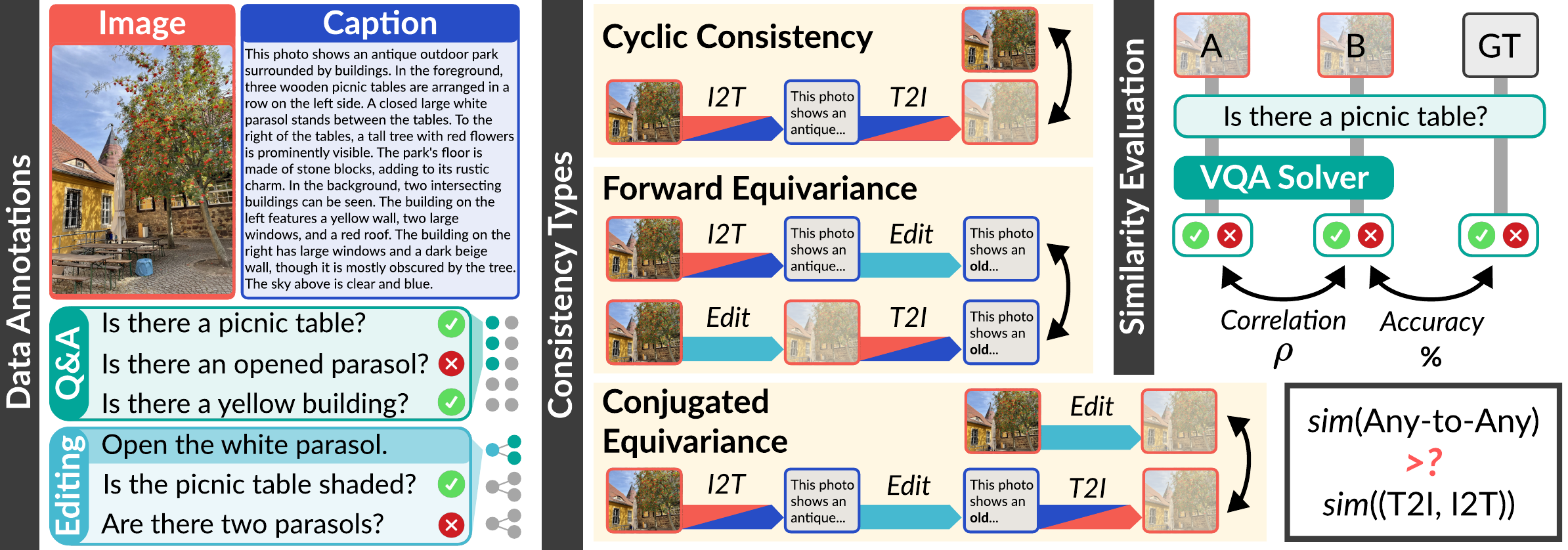}
    \caption{We evaluate whether any-to-any modality conversion models demonstrate greater consistency compared to unrelated pairs of independent image-to-text and text-to-image generators. (Left) To this end, we curate a dataset with detailed annotations, including captions, Q\&As, and editing prompts. (Centre) Consistency is measured using three criteria: cyclic consistency, forward equivariance, and conjugated equivariance, where the latter two require off-the-shelf image or text editing tools for in-modality transformations. (Right) The evaluation involves comparing the similarity between two generated outputs (images or textual descriptions) using an external VQA solver to compute correlation ($\rho$) and accuracy ($\%$).}
    \label{fig:main}
\end{figure*}

%% file: sections/method.tex
\section{Defining Consistency Across Modalities}
\label{sec:method}

We formalize the concept of multimodal consistency for numerical experiments by adopting three widely recognized types of consistency: cyclic consistency, equivariance to transformation, and commutativity of operations.

\paragraph{Notations}  
Let $\phi$ denote the parameters of a text-to-image generation model, and $\psi$ denote the parameters of an image-to-text model. We define two types of operations:  

1. Across-Modality \textit{Conversion} ($f(x)$): Transformations between modalities, such as generating an image from text ($f^{t \rightarrow i}$) or generating text from an image ($f^{i \rightarrow t}$).

2. In-Modality \textit{Modification} ($g(x, p)$): Edits or modifications within the same modality, such as image editing ($g^i$) with a given prompt $p$.
To implement $g$, we use off-the-shelf LLMs ($g^t$) and image editing models ($g^i$) because existing any-to-any models are not optimized for editing, and our focus is on evaluating cross-modality consistency rather than in-modality performance.

Furthermore, a data sample $x$ consists of two views: an image view ($x^i$) and a text view ($x^t$). If a single model is used for both modality directions, it follows that $\phi = \psi$.

\paragraph{Cyclic Consistency}
is a commonly used concept in machine learning, particularly in unpaired translation tasks~\cite{zhu2017unpaired,bielawski2023clip}. It ensures that applying transformations between two domains consecutively returns the input to its original state. For example, in unpaired image-to-image translation, translating an image from domain $A$ to domain $B$ and back to domain $A$ should reconstruct the original image.

In our setup, cyclic consistency ensures that transformations between text and image modalities are invertible. Specifically:
\begin{align}
    f_{\psi}^{i\rightarrow t}(f_{\phi}^{t\rightarrow i}(x^t)) = x^t \\
    f_{\phi}^{t\rightarrow i}(f_{\psi}^{i\rightarrow t}(x^i)) = x^i
\end{align}

\paragraph{Forward Equivariance}  
is a property that ensures the consistency of transformations under secondary operations~\cite{cohen2016group}. Specifically, applying a modification to the input followed by a transformation should yield the same result as applying the transformation first, followed by the corresponding modification. 

In this work, we adapt the traditional definition by treating modification in both modalities equivalently ($g^i \simeq g^t$).
In our context, this principle ensures compatibility between in-modality modifications and across-modality conversions:
\begin{align}
    f_{\phi}^{t \rightarrow i}(g^t(x^t, p)) = g^i(f_{\phi}^{t \rightarrow i}(x^t), p), \\
    f_{\psi}^{i \rightarrow t}(g^i(x^i, p)) = g^t(f_{\psi}^{i \rightarrow t}(x^i), p).
\end{align}
Note that forward equivariance compares transformations within the same direction, such as \( f_{\phi}^{t \rightarrow i} \) applied before or after a modification. We thus incorporate another form of equivariance relation which uses both directions at a time to evaluate consistency between modality conversions in the following paragraph.

\paragraph{Conjugated Equivariance} 
extends the idea of forward equivariance by incorporating transformations in both directions to evaluate consistency across modality conversions. Specifically, we modify forward equivariance by inverting \( f \) in the left terms, which yields:
\begin{align}
    f_{\psi}^{i\rightarrow t}(g^i(f_{\phi}^{t\rightarrow i}(x^t), p)) = g^t(x^t, p) \\
    f_{\phi}^{t\rightarrow i}(g^t(f_{\psi}^{i\rightarrow t}(x^i), p)) = g^i(x^i, p)
\end{align}
Conjugated equivariance can also be seen as an extension of cyclic consistency, where the intermediate latent space is explicitly modified before completing the transformation cycle. By incorporating multiple modifications, this approach extends pointwise evaluations to analyze structural multi-point consistency within the shared latent space.



%% file: sections/data.tex
\section{Data Collection Process}
\label{sec:data}

To support the operations outlined in~\cref{sec:method}, we curated a dataset comprising 1,000 image inputs ($i$), corresponding text captions ($t$), 3,000 editing prompts ($p$), and 16,000 question-answer pairs designed to evaluate similarity between images or captions. Further details on the data collection methodology, including human resourcing, can be found in~\cref{sec:ax_data}.

\input{figs/data}

\paragraph{Images}
The input images $i$ are curated from both \textit{unseen} and \textit{seen} sources to ensure diversity and relevance. For the \textit{unseen} subset, we collect 500 images that have not been exposed to any available MLLMs during training. Volunteers from the research community contributed private photos, which were manually filtered based on the following criteria: 1) exclusion of images with potential privacy violations or toxic content; 2) removal of images requiring domain-specific skills, such as named entity recognition or OCR capabilities; 3) elimination of low-quality images, such as those with motion blur, small resolution, or skewed aspect ratios; and 4) exclusion of overly simplistic content with very few objects. This filtering process resulted in the removal of approximately 76\% of the original submissions. For the \textit{seen} subset, we randomly sampled 500 images from the widely used COCO Captions dataset~\cite{chen2015microsoft}, which provides a benchmark set of images commonly utilized in multimodal learning.

\paragraph{Captions}  
The goal of our captioning process is to create captions that accurately guide the reconstruction of the original image. We manually annotate dense captions \( t \) for input images \( i \), employing the communication game framework~\cite{kim2019codraw} to ensure alignment between modalities. Annotators are assigned three roles: the \textit{teller}, the \textit{drawer}, and the \textit{judge}. 

The teller crafts a caption \( t \) that encodes the essential visual details of the image \( i \), acting as the sender in the communication game. The drawer interprets the caption and reconstructs the image using tool-assisted generation (DALL·E-3~\cite{betker2023improving} and Imagen 2~\cite{imagen2}), functioning as the receiver. The reconstructed image \( \hat{i} \) is compared to the original image \( i \) to evaluate the success of the communication. To maintain objectivity, the teller does not have access to the reconstructed image during annotation, ensuring that captions are crafted independently of the reconstruction process. Finally, the \textit{judge} evaluates the quality of the reconstructed image and provides feedback (good/bad). If necessary, re-captioning is requested from a different \textit{teller}.

\paragraph{Questions \& Answers}  
A reliable metric is essential to assess factual similarity between images. Retrieval-based metrics, such as CLIPScore~\cite{hessel2021clipscore}, are insufficient for evaluating factual correctness, particularly in compositionality~\cite{ma2023crepe} or counting tasks~\cite{radford2021learning}. Recent studies~\cite{hu2023tifa,chodavidsonian} propose an alternative approach: generating questions that capture salient facts about the images. One such metric, VQAScore~\cite{lin2024evaluating}, automatically generates questions using a pretrained VQA question generator and then scores consistency by comparing model answers on reference and generated images. While VQAScore improves over retrieval-based methods by grounding evaluation in factual content, its reliability is limited by the quality and scope of automatically generated questions, which tend to be shallow and generic, failing to stress fine-grained or compositional differences.

By contrast, our approach introduces a human-in-the-loop \textit{comparer}, who carefully constructs challenging questions that highlight both similarities and differences between the reference and reconstructed images. These questions are specifically designed to distinguish subtle failures in object placement, count, or relational semantics that automatic systems often overlook. For each image pair, five similarity- and five difference-oriented questions are created, ensuring coverage of both aligned and misaligned aspects. This deliberate design leads to more sensitive and discriminative factual evaluations than automatic pipelines such as VQAScore.

\paragraph{Editing Operations}  
To evaluate equivariance and commutativity properties, we define modification operations for each modality (\( g^i \) for images and \( g^t \) for text). These operations are conditioned on a prompt \( p \), specifying the nature and direction of the modification. The \textit{comparer} annotates three prompts per image, ensuring alignment with the intended changes, and additionally generates two prompt-conditioned question-answer pairs per editing prompt to reflect the specific modifications.

\paragraph{Manual Filtering}
To address quality variance inherent in collaborative annotation, we conducted a rigorous manual filtering process aimed at normalizing differences across annotators. Observed inconsistencies included variation in the verbosity of image descriptions, factual correctness, and formatting conventions in Q\&As (e.g., inconsistent use of parentheses to denote objects or attributes). To ensure consistency, we established strict filtering criteria: (1) questions must be answerable based solely on the provided description, (2) answers must be factually correct, (3) the question set must be diverse and cover different object types or visual properties, and (4) descriptions must contain sufficient detail. Any factual inconsistencies led to automatic rejection.

Before initiating the filtering process, all reviewers (distinct from the original annotators) participated in a calibration phase to align evaluation standards. Each reviewer shared and critiqued 10 annotated examples with others, enabling discussion on interpretation and enforcement of the filtering criteria. After normalization, each sample was independently reviewed by two human judges. Approximately 43\% of the initial samples were discarded and replaced with new annotations that met all quality standards.


%% file: figs/data.tex
\begin{figure}
    \centering
    \includegraphics[width=0.48\textwidth]{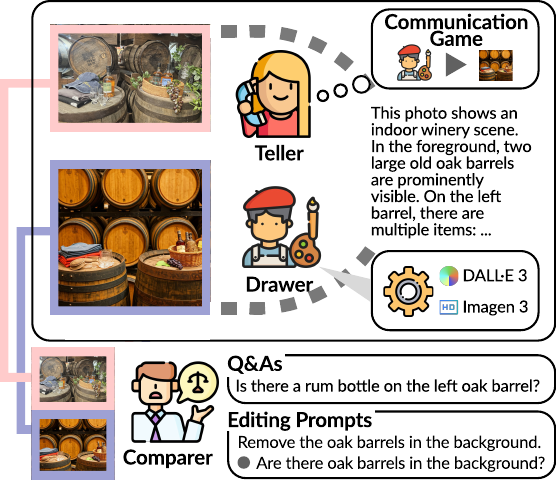}
    \caption{
    Data annotation process for \modelname. Three human workers perform distinct roles: the \textit{teller} creates a textual description emphasizing key elements for reconstruction, following a communication game framework~\cite{kim2019codraw}, the \textit{drawer} recreates the image using the description via multi-turn AI interactions, and the \textit{comparer} generates Q\&As to capture similarities and differences between the original and reconstructed images, annotating editing instructions as well.
    }
    \label{fig:data_collection}
\end{figure}

%% file: sections/implementation.tex

%% file: sections/experiments.tex
\section{Experiments}
\label{sec:experiments}

\subsection{Setups}
\label{subsec:setup}

This section outlines the models tested and the utilities employed in our experiments. Full details, including model checkpoints and instruction prompts, are available in~\cref{sec:ax_setups} and~\cref{sec:ax_prompts}.

\paragraph{Models}  
For \textit{text-to-image} generation, we use the open-source models including Flux~\cite{blackforest2024flux} and Stable Diffusion XL~\cite{podellsdxl}. These models were selected for their balance between performance and resource efficiency, allowing the evaluation to focus on consistency rather than absolute image fidelity. For \textit{image-to-text} generation, we include LLaVA-Next~\cite{liu2024improved} (abbreviated as LLaVA) and Qwen2VL~\cite{bai2023qwen}. These models are chosen for their widespread use, robust performance across tasks, and practical trade-offs in computational requirements. To assess the central claim that \textit{any-to-any models} improve cross-modal coherence, we evaluate four open-source systems: Chameleon~\cite{team2024chameleon}, Emu-3~\cite{wang2024emu3}, VILA-U~\cite{wu2024vila}, and Seed-X~\cite{ge2024seed}. For model descriptions, refer to~\cref{sec:related_work}.

\paragraph{Utilities}  
\textit{In-Modality Editors}: For editing within a single modality, we use Cos Stable Diffusion XL 1.0 Edit (CosXL)~\cite{podellsdxl} for image editing and Qwen2.5~\cite{yang2024qwen2} for text editing. \textit{Evaluators}: For visual question-answering tasks, we employ PaliGemma2~\cite{steiner2024paligemma} for its strong performance in static VQA scenarios. We use Qwen2.5 for textual Q\&As.

\input{tables/cycle}

\subsection{Cyclic Consistency}
\label{subsec:exp_cycle}

\textit{Cyclic consistency} refers to the model's capability to accurately reconstruct input data by leveraging its latent representations, ensuring the preservation of original content through a bidirectional transformation process. For image reconstruction, the process involves an image-to-text transformation followed by a text-to-image transformation ($x \xrightarrow{f^{i \rightarrow t}} z \xrightarrow{f^{t \rightarrow i}} \bar{x}$). For text reconstruction, the order is reversed ($x \xrightarrow{f^{t \rightarrow i}} z \xrightarrow{f^{i \rightarrow t}} \bar{x}$). The reconstructed data $\bar{x}$ is compared to the original input $x$ for evaluation.

\paragraph{Metrics}
We employ off-the-shelf visual or textual question-answering tools to compare the reconstructed data $\bar{x}$ with the original input $x \in \mathcal{X}$. Given the model output $\bar{x}$, the context $c \in \mathcal{C}$, and a question $q \in \mathcal{Q}$, evaluation is conducted using a parameterized binary classifier $h_\theta: \mathcal{X} \times \mathcal{C} \times \mathcal{Q} \to \{0, 1\}$. The primary accuracy metric is defined as:
\begin{align}
    sim(x, \bar{x}) := \sum_{q} \delta \big( h_\theta (\bar{x}, c, q), h_{o}(x, c, q) \big),
\end{align}
where $\delta$ is the Dirac delta function, and $h_{o}$ represents the oracle classifier that provides the ground-truth labels. 
Note that we average over ten different questions per model-generated output.
In addition to accuracy, we report the F1-score, which incorporates precision and recall, to assess the similarity between the binary outputs.

\paragraph{Results}
The cyclic consistency evaluation results are presented in~\cref{tab:combined_cycle}. Notably, a single any-to-any model does not consistently outperform the combination of separate specialized models in cyclic consistency, raising questions about the presumed advantages of training a single any-to-any model.

Notable exceptions include Seed-X and VILA-U, which demonstrate notable consistency when utilizing a single any-to-any model. In contrast, other any-to-any models such as Chameleon and Emu3 fail to exhibit consistent patterns. This disparity aligns with the visual tokenization strategies employed by these models: both Seed-X and VILA-U adopt \textit{semantically-aligned} visual tokenizers, either by leveraging features from a pre-trained ViT or by optimizing alignment with textual representations. On the other hand, Chameleon and Emu3 rely solely on image reconstruction objectives.
This finding indicates that incorporating semantic modeling into visual tokenization may contribute to improved alignment of the latent space during modality conversions.

Still, the results show that this evaluation conflates per-modality transfer performance with cyclic consistency. For instance, VILA-U, when used as the secondary text-to-image operator, achieves high performance regardless of the initial operation it is paired with. A similar trend is observed in text generation, where models such as LLaVA and Qwen2VL tend to outperform others.
In conclusion, evaluating any-to-any consistency requires multiple complementary criteria, which we address in the following experiments.

\subsection{Forward Equivariance}
\label{subsec:exp_equi}

\textit{Forward Equivariance} assesses the impact of applying an in-modality editing operation (\( g \)) either before or after the modality transfer (\( f \)). Unlike other consistency criteria, this approach focuses on comparing outputs from the same modality transfer direction (\( f^{i \rightarrow t} \) vs. \( f^{i \rightarrow t} \) and \( f^{t \rightarrow i} \) vs. \( f^{t \rightarrow i} \)).

\paragraph{Metrics}
This evaluation involves three key comparisons: the two model outputs (\( f(g(x)) \) and \( g(f(x)) \)) and the ground-truth modified datapoint \( x' \). The primary metric is the Pearson correlation between \( f(g(x)) \) and \( g(f(x)) \), emphasizing consistency over absolute performance. Additional metrics, \( sim(f(g(x)), x') \) and \( sim(g(f(x)), x') \), are detailed in the appendix.

Similarity between datapoints is measured using question-answering methods, as in the cyclic consistency evaluation. However, each question and answer here is conditioned on an editing prompt \( p \in \mathcal{P} \). For each sample, we compute averages over two prompt-conditioned questions per editing prompt, using three editing prompts per image or text.

\paragraph{Results}
We illustrate correlation statistics in~\cref{fig:equi}, while the complete results are presented in~\cref{sec:ax_exp}. The findings reaffirm earlier observations: the consistency of any-to-any models relative to independent specialist pairs is not consistently superior. However, notable exceptions include Seed-X and VILA-U, which exhibit improved textual consistency, consistent with trends observed in previous experiments.

\input{figs/equi}
\input{figs/symmetry}

\subsection{Conjugated Equivariance}
\label{subsec:exp_sym}

\textit{Conjugated Equivariance} extends cyclic consistency by applying an in-modality operation $g$ between modality transfers. Taking image reconstruction as an example, the goal is to reconstruct the image with the modification in the latent textual description represented correctly ($x \xrightarrow{f^{i \rightarrow t}} z \xrightarrow{g^t} z' \xrightarrow{f^{t \rightarrow i}} \bar{x}'$).
This approach shifts the focus from evaluating single-point reconstructions to assessing the alignment of transformation directions (vectors) across modalities. The process is applied analogously for textual reconstruction.

\paragraph{Metrics}
This evaluation compares two terms: the model output $\bar{x}'$ and the ground-truth label $x'$. Thus, we report accuracy and F1 score as in the cyclic consistency experiment. The only difference is that the questions are also conditioned on the editing prompts. Thus, we average results over two questions per editing prompt, testing three editing prompts per sample.
We do not generate the in-modality output ($g(x, p)$). Instead, we use ground-truth answers to the question to replace the evaluation results  ($h_o(g(x, p), c, p)$).

\paragraph{Results}
Empirical results, visualized in~\cref{fig:symmetry} and detailed further in~\cref{sec:ax_exp}, reveal consistent self-alignment for most any-to-any models. All any-to-any models, except for Chameleon in image generation, achieve stable self-consistency when paired with themselves. However, these models do not consistently outperform when paired with themselves compared to being paired with other models. This suggests that while any-to-any models demonstrate stable self-alignment, they are not always the optimal choice for their own outputs in cross-modal operations.

This raises the question: why is any-to-any consistency, barely noticeable in cyclic consistency, more evident in the conjugated equivariance experiment? A plausible explanation is that conjugated equivariance evaluates consistency by analyzing transformations across a distribution of edited latent representations, rather than focusing on a single transformation. By leveraging multiple editing operations, this approach captures broader patterns of alignment in the latent space, enabling a more nuanced assessment of consistency between text-to-image and image-to-text capabilities. This finding aligns with the shared latent learning hypothesis~\cite{huang2021makes,lu2023theory}, which posits that models trained on multiple tasks or modalities form a unified latent space for shared representations.

\input{figs/inference}

\subsection{Qualitative Results}
\label{subsec:exp_qual}

Figure~\ref{fig:inference} presents sample inference results used to evaluate cyclic consistency and conjugated equivariance. A key observation is the stylistic disparity between natural (ground truth) and model-generated images, underscoring the limitations of cyclic consistency as a reliability metric. This supports the use of equivariance, which directly compares outputs across models and avoids distortions arising from differences between natural and synthetic images. Furthermore, Chameleon's poor image fidelity often aligns with object composition errors (e.g., incorrect counts or misplaced elements), highlighting that any-to-any models do not consistently translate their stronger linguistic capabilities into accurate compositional image generation.

\subsection{Discussion}

\paragraph{Diversity}
This work does not explicitly address the diversity of generated images or text, as we employ deterministic sampling throughout. While distributional analysis would be ideal for evaluating coverage of semantic space, its effectiveness is constrained by the limited generative diversity of current image synthesis models. As noted in prior work~\cite{hsieh2024graph}, models such as Stable Diffusion XL~\cite{podellsdxl} often fail to produce semantically distinct outputs even when conditioned on different random seeds, limiting the effectiveness of stochastic sampling for diversity evaluation.

%% file: tables/cycle.tex
\begin{table*}[h!]
\small
\centering
\begin{tabular}{llcc|llcc}
\toprule
\multicolumn{4}{c|}{\textbf{Image $\rightarrow$ Text $\rightarrow$ Image}} & \multicolumn{4}{c}{\textbf{Text $\rightarrow$ Image $\rightarrow$ Text}} \\
\textbf{I2T} & \textbf{T2I} & \textbf{Accuracy (\%)} & \textbf{F1 (\%)} & \textbf{T2I} & \textbf{I2T} & \textbf{Accuracy (\%)} & \textbf{F1 (\%)} \\ \toprule
\multirow{6}{*}{LLaVA} & Flux & \underline{61.78} & \underline{72.29} & \multirow{6}{*}{Flux} & LLaVA & \underline{63.73} & \underline{70.28} \\
 & SDXL & 57.21 & 67.91 &  & Qwen2VL &  \textbf{66.93} &  \textbf{73.61} \\
 & Chameleon & 56.13 & 66.77 &  & Chameleon & 56.42 & 61.95 \\
 & Emu3 & 60.91 & 71.65 &  & Emu3 & 63.48 & 70.03 \\
 & Seed-X & \textbf{62.57} & \textbf{73.37} &  & Seed-X & 63.52 & 70.22 \\
 & VILA-U & 61.40 & 71.99 &  & VILA-U & 62.36 & 68.59 \\\midrule
\multirow{6}{*}{Qwen2VL} & Flux & \underline{62.86} & 73.20 & \multirow{6}{*}{SDXL} & LLaVA & \underline{59.20} & \underline{65.10} \\
 & SDXL & 57.83 & 68.53 &  & Qwen2VL & 58.04 & 63.98 \\
 & Chameleon & 57.04 & 67.65 &  & Chameleon & 57.17 & 62.63 \\
 & Emu3 & 61.50 & 73.28 &  & Emu3 & 57.58 & 63.07 \\
 & Seed-X & 62.65 & \underline{73.33} &  & Seed-X & 57.29 & 62.67 \\
 & VILA-U & \textbf{63.36} & \textbf{73.92} &  & VILA-U & \textbf{59.70} & \textbf{65.43} \\\midrule
\multirow{6}{*}{Chameleon} & Flux & 55.38 & 65.75 & \multirow{6}{*}{Chameleon} & LLaVA & \underline{55.63} & \textbf{60.82} \\
 & SDXL & 54.47 & 64.85 &  & Qwen2VL & 54.38 & 59.06 \\
 & \textcolor{paircolor}{Chameleon} & \textcolor{paircolor}{53.93} & \textcolor{paircolor}{64.33} &  & \textcolor{paircolor}{Chameleon} & \textbf{\textcolor{paircolor}{55.76}}  & \underline{\textcolor{paircolor}{60.59}}  \\
 & Emu3 & 55.81 & 66.23 &  & Emu3 & 53.97 & 58.66 \\
 & Seed-X & \underline{57.04} & \underline{68.05} &  & Seed-X & 54.59 & 59.32 \\
 & VILA-U & \textbf{59.04} & \textbf{69.79} &  & VILA-U & 55.46 & 60.24 \\\midrule
\multirow{6}{*}{Emu3} & Flux & 58.00 & 68.63 & \multirow{6}{*}{Emu3} & LLaVA & \textbf{62.61} & \textbf{68.75} \\
 & SDXL & 54.01 & 64.28 &  & Qwen2VL & \underline{61.40} & \underline{67.75} \\
 & Chameleon & 53.72 & 63.85 &  & Chameleon & 57.46 & 63.06 \\
 & \textcolor{paircolor}{Emu3} & \textcolor{paircolor}{58.62} & \textcolor{paircolor}{69.20} &  & \textcolor{paircolor}{Emu3} & \textcolor{paircolor}{60.03} & \textcolor{paircolor}{66.17} \\
 & Seed-X & \underline{59.04} & \underline{69.79} &  & Seed-X & 61.15 & 67.23 \\
 & VILA-U & \textbf{59.29} & \textbf{69.98} &  & VILA-U & 59.20 & 64.93 \\\midrule
\multirow{6}{*}{Seed-X} & Flux & 60.91 & 71.39 & \multirow{6}{*}{Seed-X} & LLaVA & 59.95 & 65.79 \\
 & SDXL & 57.46 & 67.94 &  & Qwen2VL & \underline{60.95} & \underline{67.32} \\
 & Chameleon & 56.13 & 66.54 &  & Chameleon & 56.42 & 61.76 \\
 & Emu3 & 60.53 & 71.28 &  & Emu3 & 57.46 & 63.09 \\
 & \textcolor{paircolor}{Seed-X} & \textbf{\textcolor{paircolor}{61.78}} & \textbf{\textcolor{paircolor}{72.53}} &  & \textcolor{paircolor}{Seed-X} & \textbf{\textcolor{paircolor}{61.45}} & \textbf{\textcolor{paircolor}{67.61}} \\
 & VILA-U & \underline{61.07} & \underline{71.65} &  & VILA-U & 59.41 & 65.12 \\\midrule
\multirow{6}{*}{VILA-U} & Flux & 60.41 & 70.97 & \multirow{6}{*}{VILA-U} & LLaVA & 55.84 & 60.99 \\
 & SDXL & 58.37 & 69.00 &  & Qwen2VL & \underline{57.87} & \underline{63.47} \\
 & Chameleon & 55.13 & 65.52 &  & Chameleon & 56.25 & 61.53 \\
 & Emu3 & \underline{60.82} & \underline{71.66} &  & Emu3 & 55.46 & 60.50 \\
 & Seed-X & 60.70 & 71.64 &  & Seed-X & 56.50 & 61.94 \\
 & \textcolor{paircolor}{VILA-U} & \textbf{\textcolor{paircolor}{62.15}} & \textbf{\textcolor{paircolor}{72.88}} &  & \textcolor{paircolor}{VILA-U} & \textbf{\textcolor{paircolor}{58.12}} & \textbf{\textcolor{paircolor}{63.69}} \\\bottomrule

\end{tabular}
\caption{Cyclic consistency evaluation results. The best scores for the same initial operation are highlighted in \textbf{bold}, while the second-best scores are \underline{underlined}. Results obtained using a single any-to-any model, instead of distinct model pairs, are presented in \textcolor{paircolor}{color}.}
\label{tab:combined_cycle}
\end{table*}

%% file: figs/equi.tex
\begin{figure}[!ht]
    \centering
    \includegraphics[width=0.48\textwidth]{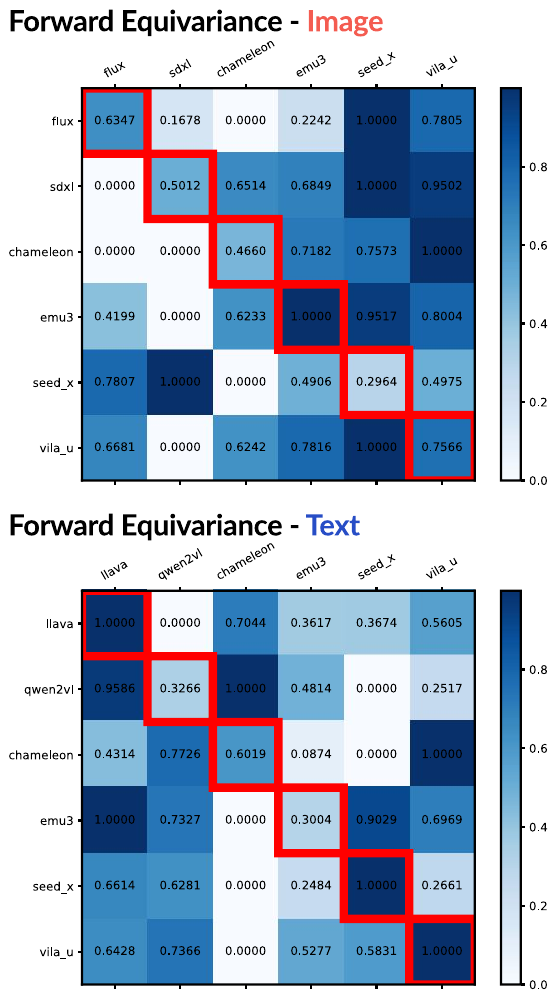}
    \caption{\textit{Correlation} of forward equivariance across model pairs, normalized to [0, 1] per row. Diagonal components with red borders indicate the same any-to-any generator used for both image-to-text and text-to-image transfers.}
    \label{fig:equi}
\end{figure}

%% file: figs/symmetry.tex
\begin{figure}[!ht]
    \centering
    \includegraphics[width=0.48\textwidth]{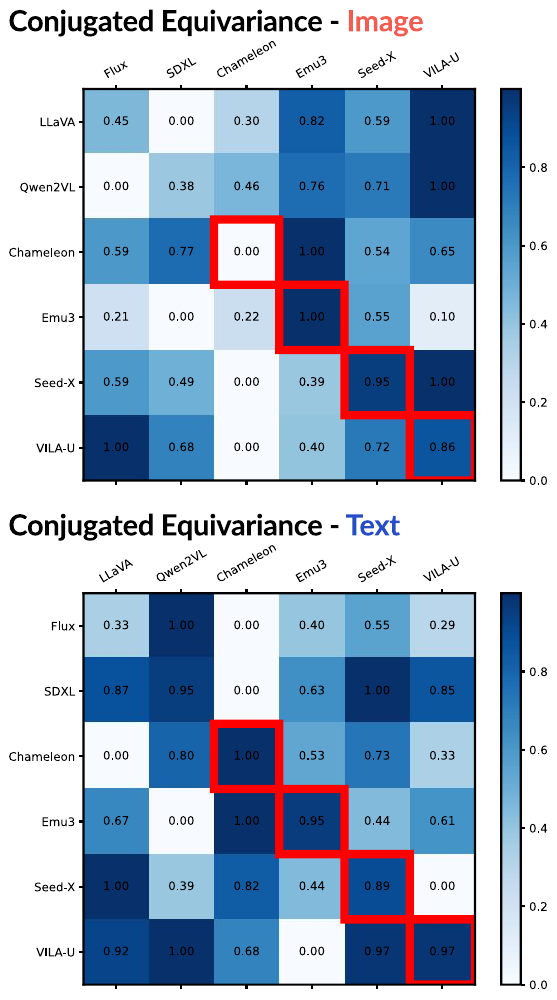}
    \caption{\textit{Accuracy} of conjugated equivariance across model pairs, normalized to [0, 1] per row. Diagonal components with red borders indicate the same any-to-any generator used for both image-to-text and text-to-image transfers.}
    \label{fig:symmetry}
\end{figure}

%% file: figs/inference.tex
\begin{figure*}[!ht]
    \centering
    \includegraphics[width=0.98\textwidth]{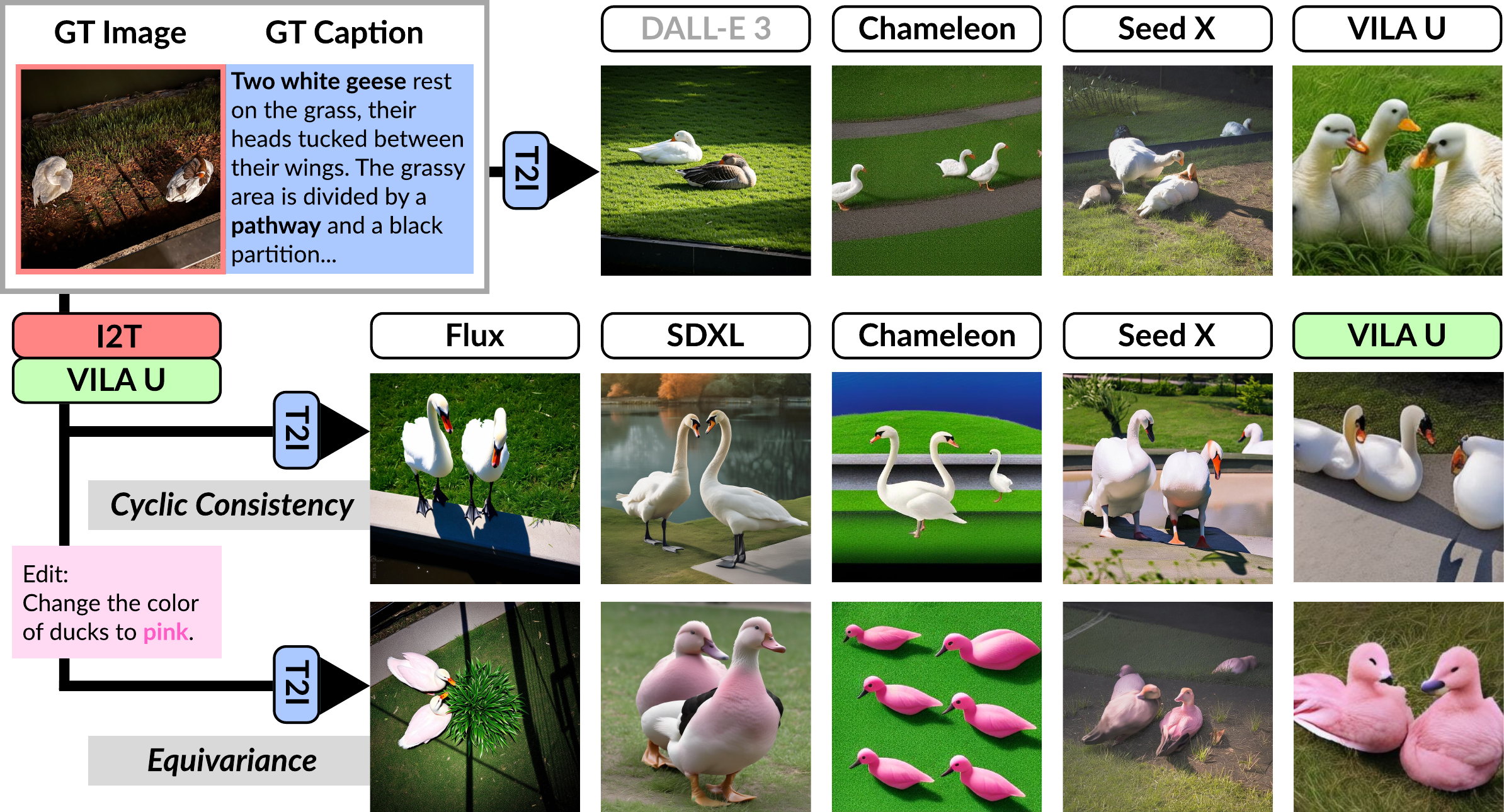}
    \caption{Example inference results using VILA-U~\cite{wu2024vila} as an example image-to-text captioner. (Top row) Text-to-image transfer results are shown using the human-annotated caption as input for reference. (Middle row) \textit{Cyclic consistency} in the image domain is evaluated by captioning the original image with the captioner, then reconstructing it using different image generators. (Bottom row) \textit{Conjugated Equivariance} is assessed by applying an editing operation $g$ in the latent domain before reconstructing the image.}
    \label{fig:inference}
\end{figure*}

%% file: sections/related_work.tex
\section{Related Work}
\label{sec:related_work}




\paragraph{Any-to-Any Models}
Any-to-any generative models aim to unify multimodal understanding and generation across diverse tasks and modalities. Here, we focus on image and text modalities to align with the scope of this paper. 
Recent approaches can be categorized into deterministic and distributional modeling of image data. Deterministic approaches, such as Kosmos-G~\cite{pan2024kosmosg} and Emu2~\cite{sun2024generative}, directly regress CLIP~\cite{radford2021learning} features, which are then fed into an (optionally fine-tuned) Stable Diffusion~\cite{rombach2022stable} generator. Distributional approaches, by contrast, often compress images into discrete token sequences using vector quantization~\cite{oord2017vqvae}, enabling categorical latent space modeling. Examples include OFA~\cite{wang2022ofa}, Unified-IO 2~\cite{lu2024unified}, Chameleon~\cite{team2024chameleon}, LaVIT~\cite{jin2023unified}, and Emu3~\cite{wang2024emu3}.
To enhance information sharing between text-to-image and image-to-text tasks, recent models such as SEED-LLaMA~\cite{ge2024making}, SEED-X~\cite{ge2024seed}, and VILA-U~\cite{wu2024vila} incorporate semantic alignment into their tokenization strategies. Additionally, diffusion-based approaches, exemplified by Transfusion~\cite{zhou2024transfusion}, are emerging as alternatives to categorical tokenization, leveraging continuous distributions for greater flexibility.

\paragraph{Multimodal Consistency}
Multimodal consistency ensures coherence across modalities. MM-R3~\cite{chou2024mm} and MMCBench~\cite{zhang2024benchmarking} evaluate robustness to semantic shifts and corrupted inputs. Advances in text-to-image consistency include PDF-GAN~\cite{tan2022ssd}, which employs Semantic Similarity Distance, and a diffusion framework~\cite{sun2024prompt} leveraging knowledge graphs. MC-MKE~\cite{zhang2024mc} addresses modality errors, while ConsiStory~\cite{tewel2024training} improves layout consistency without additional training. CycleGAN~\cite{bielawski2023clip} and CyclePrompt~\cite{diesendruck2024learning} enhance captioning and code generation with cycle-supervised methods. Semantic consistency metrics~\cite{bent2024semantic} and cycle-consistency losses~\cite{zhu2017unpaired} further refine reliability.


%% file: sections/conclusion.tex
\section{Conclusion}
\label{sec:conclusion}
We introduce \modelnamefancy, a hand-annotated benchmark designed to evaluate the any-to-any consistency of multimodal AI models. Our analysis reveals that existing any-to-any models exhibit weak consistency between text-to-image and image-to-text tasks, which becomes apparent only through distributional inspections of the intermediate latent space, facilitated by multiple editing operations.

\section{Limitations \& Future Directions}
\label{sec:limitations}

\paragraph{Limitations}
Our experiments are conducted using any-to-any models in their \textit{as-is} state. Since model behavior results from the interplay of data, architecture, and training processes, this monolithic evaluation does not allow us to isolate the specific factors contributing to (in)consistency across modalities. We encourage the research community, particularly those with greater computational resources, to undertake controlled analyses to systematically examine how each design component of any-to-any models impacts their consistency.

Our new benchmark, \modelname, has certain limitations stemming from its curation process, which focuses on hand-taken natural images. This emphasis impacts the dataset's image distribution in several ways. First, our private images exclude artistic images, 2D drawings, and 3D renderings, limiting the scope for evaluating any-to-any consistency in these domains. Second, as the dataset relies on pre-taken image contributions, the subjects are predominantly confined to realistic scenarios typically captured by people, such as scenic landscapes, food, or animals. Although efforts were made to ensure diversity, these inherent distributional biases persist in the dataset.

Additionally, \modelname was annotated by five NLP researchers sharing similar cultural backgrounds. Although a separate group of human evaluators validated these annotations, we acknowledge the potential influence of cultural bias on image descriptions.
For instance, studies~\cite{nisbett2001culture,ananthram2024see} suggest that individuals from Western cultures often emphasize the central figure in an image, while those from Eastern cultures are more inclined to consider the broader scene context.

\paragraph{Future Directions}

Future directions include:
\begin{enumerate}
    \item \textbf{Iterative Composition}: This work focuses on a single cyclic loop of modality transfers. Exploring iterative composition of transfers could provide further insights into consistency. Neural networks approximating data distributions are known to collapse output diversity under repeated application—would consistent cyclic loops mitigate this effect?
    \item \textbf{Extending Modalities}: Expanding beyond the (image, text) modality pair to others, such as (speech, text), could uncover whether any-to-any models demonstrate stronger consistency across different domains.
\end{enumerate}

\paragraph{Risks}  
We introduce a new multimodal data corpus, including newly contributed private images. Each image has undergone manual inspection to prevent copyright infringement, portrait rights violations, and the inclusion of harmful or inappropriate content. However, some risks remain:

\begin{itemize}
    \item \textbf{Bias and Representational Gaps}: Despite efforts to ensure diversity, the dataset may inadvertently overrepresent or underrepresent certain cultural or demographic backgrounds, potentially leading to biased model outputs or unfair generalizations.
    \item \textbf{Unintended Personal Data Exposure}: While we obtained explicit consent from contributors and filtered out any images that could reveal their identity, advancements in AI, such as geographic inference models, may enable the extraction of private information from images in unintended and non-explicit ways.
    \item \textbf{Erosion of Zero-Shot Integrity}: By releasing new private images, we aim to encourage evaluation on truly unseen data. However, public availability of the dataset risks its use for fine-tuning future models, potentially compromising the integrity of results in subsequent zero-shot evaluations.
\end{itemize}

%% file: sections/ack.tex
\section{Acknowledgements}
\label{sec:acknowledgement}

This work was supported by Institute of Information \& communications Technology Planning \& Evaluation (IITP) grant funded by the Korea government(MSIT) (No.RS-2024-00457882, AI Research Hub Project; No.RS-2020-II201361, Artificial Intelligence Graduate School Program (Yonsei University); and No.RS-2025-02263598, Development of Self-Evolving Embodied AGI Platform Technology through Real-World Experience)
and the National Research Foundation of Korea (NRF) grant funded by the Korea government (MSIT) (No. RS-2024-00354218).

%% file: sections/ax_setups.tex
\section{Experimental Setups}
\label{sec:ax_setups}

\paragraph{Model Versions}
We use official versions from HuggingFace, except for the Anole~\cite{chern2024anole} variant of Chameleon, selected for its acceptable image quality. Detailed information is provided in~\cref{table:ax_models}.

\input{tables/ax_models}

\paragraph{Implementations}
We integrate the official inference codes for each model into a unified codebase. All hyperparameters and pipelines are preserved as originally provided, with the addition of a wrapper to standardize the interface and dependencies. Detailed implementation references are available in our codebase. Below are the original codebases for models that require specific implementations beyond the standard \texttt{transformers} or \texttt{diffusers} pipelines:
\begin{itemize}
\item \textbf{Anole}: \url{https://github.com/GAIR-NLP/anole}
\item \textbf{Emu3}: \url{https://github.com/baaivision/Emu3}
\item \textbf{VILA-U}: \url{https://github.com/mit-han-lab/vila-u}
\item \textbf{Seed-X}: \url{https://github.com/AILab-CVC/SEED-X}
\end{itemize}

\paragraph{Stochasticity}
Our experiments are predominantly deterministic, as we perform inference-only evaluations, with most models recommending greedy decoding over random sampling for optimal results. Exceptions include diffusion~\cite{podellsdxl} and flow-matching~\cite{blackforest2024flux} models (e.g., SDXL and Flux), which inherently involve stochasticity due to latent sampling. To maintain consistency, we run these models a single time, aligning them with the deterministic models.

\paragraph{Computation}
All experiments were conducted using 8 NVIDIA L40S GPUs, each with 48GB VRAM. Multi-GPU parallelization was not utilized, as each model tested fits within a single L40S GPU when using half-precision and a batch size of one.

%% file: tables/ax_models.tex
\begin{table*}[t!]
\centering
\resizebox{0.98\textwidth}{!}{
\begin{tabular}{l|lcl}
\toprule
Type                   & Name   & Size & Backbone \\ \hline
Text-to-Image & Flux~\cite{blackforest2024flux} & 12B & \texttt{black-forest-labs/FLUX.1-schnell} \\
& SDXL~\cite{podellsdxl}  &  3.5B & \texttt{stabilityai/stable-diffusion-xl-base-1.0} \\
Image-to-Text& LLaVA-Next~\cite{liu2024improved}  & 8B & \texttt{llava-hf/llama3-llava-next-8b-hf} \\
& Qwen2VL~\cite{bai2023qwen} & 7B & \texttt{Qwen/Qwen2-VL-7B-Instruct} \\
Any-to-Any& Chameleon~\cite{team2024chameleon}  & 7B & \texttt{GAIR/Anole-7b-v0.1}~\cite{chern2024anole} \\
& Emu3~\cite{wang2024emu3} & 8B & \texttt{BAAI/Emu3-Stage1} \\
& VILA-U~\cite{wu2024vila} & 7B & \texttt{mit-han-lab/vila-u-7b-256} \\
& Seed-X~\cite{ge2024seed} & 17B & \texttt{AILab-CVC/seed-x-17b-instruct} \\
\midrule
Editor & CosXL~\cite{podellsdxl} & 3.5B & \texttt{stabilityai/cosxl} \\
& Qwen2.5~\cite{yang2024qwen2} & 7B & \texttt{Qwen/Qwen2.5-7B-Instruct} \\
\midrule
Evaluator & PaliGemma2~\cite{steiner2024paligemma} & 10B & \texttt{google/paligemma2-10b-pt-448} \\
& Qwen2.5~\cite{yang2024qwen2} & 7B & \texttt{Qwen/Qwen2.5-7B-Instruct} \\
\bottomrule
\end{tabular}}
\caption{Specifications of the models used in our experiments, categorized by their functionality. The table lists the type of operation (e.g., Text-to-Image, Image-to-Text, Any-to-Any), model names, parameter sizes, and corresponding backbones. Editors and evaluators used for in-modality editing and cross-modal evaluations are also included.}
\label{table:ax_models}
\end{table*}

%% file: sections/ax_data.tex
\section{Data Annotation}
\label{sec:ax_data}

\paragraph{Human Resources}
The authors manually annotated the corpus, involving four distinct roles. The \textit{teller} viewed the image and created captions, while the \textit{drawer} reconstructed the image through multi-turn interactions with proprietary AI image generation models. The \textit{comparer} assessed the original and reconstructed images, annotating Q\&A pairs along with editing instructions. Finally, a fourth individual reviewed each annotation for validity. Annotations marked as invalid were discarded, and the corresponding image was re-annotated. On average, each image underwent $~\sim2.12$ annotation cycles to meet the high-quality standards.

\paragraph{Private Image Contribution}
Images were privately contributed by colleagues within our organization. To prevent potential copyright issues, only images taken by the contributors themselves were accepted. Contributed images were manually filtered based on the following criteria:
\begin{itemize}
\item The image must not contain harmful content.
\item The image must not contain content that can reveal the contributor's identity.
\item The image must include at least two objects.
\item The image must not contain excessive scene text.
\item The resolution must meet or exceed $1024 \times 1024$ pixels.
\end{itemize}
From approximately 2,000 contributed images, 500 were retained for inclusion in the corpus after applying these filters.

\paragraph{Caption Annotation}
To ensure consistency across annotators, we provide the following guidelines for human annotators (\textit{tellers}) during caption generation:

\begin{itemize} \item Captions should be optimized to facilitate accurate reconstruction of the original image. \item Each caption should be approximately a paragraph in length ($\sim$7–10 sentences). \item Captions must avoid affective expressions and exclude inferred commonsense facts not directly observable in the image. \item Each sentence should be self-contained, refraining from referencing relative positions (e.g., "to its right," "to his left," "behind them"). \item Still Life: Describe elements beginning from the center and progressing outward. \item Landscape: Describe elements sequentially from the foreground to the background (foreground → midground → background). \end{itemize}

%% file: sections/ax_exp.tex
\section{Additional Experimental Results}
\label{sec:ax_exp}

\input{tables/equiv}

\paragraph{Forward Equivariance}
Detailed numerical results for the experiment summarized as a confusion matrix in~\cref{subsec:exp_equi} of the main paper are provided here. Refer to~\cref{tab:equiv} for the complete results.

\input{tables/sym}

\paragraph{Conjugated Equivariance}
This section includes the detailed numerical results for the conjugated equivariance consistency test. The results analyze how well the models preserve symmetry in transformations, as outlined in~\cref{subsec:exp_sym} of the main paper. The full table of results is provided in~\cref{tab:sym}.

%% file: tables/equiv.tex
\begin{table*}[h!]
\resizebox{0.98\textwidth}{!}{
\centering
\begin{tabular}{llccc|llccc}
\toprule
\multicolumn{5}{c|}{\textbf{Image}} & \multicolumn{5}{c}{\textbf{Text}} \\
\textbf{I2T} & \textbf{T2I} & \textbf{Acc $f(g(x))$} & \textbf{Acc $g(f(x))$} & \textbf{Corr} & \textbf{I2T} & \textbf{T2I} & \textbf{Acc $f(g(x))$} & \textbf{Acc $g(f(x))$} & \textbf{Corr} \\ \toprule 
\multirow{6}{*}{Flux} & \textcolor{paircolor}{Flux} & \textcolor{paircolor}{61.92} & \textcolor{paircolor}{44.77} &  \textcolor{paircolor}{0.3482} & \multirow{6}{*}{LLaVA} & \textcolor{paircolor}{LLaVA} & \textcolor{paircolor}{48.10} & \textcolor{paircolor}{62.15} &  \textcolor{paircolor}{0.2401} \\
 & SDXL & 61.92 & 44.77 & 0.3122 &  & Qwen2VL & 48.10 & 62.15 & 0.0282 \\
 & Chameleon & 61.92 & 44.77 & 0.2993 &  & Chameleon & 48.10 & 62.15 & 0.1774 \\
 & Emu3 & 61.92 & 44.77 & 0.3166 &  & Emu3 & 48.10 & 62.15 & 0.1048 \\
 & Seed-X & 61.92 & 44.77 & 0.3764 &  & Seed-X & 48.10 & 62.15 & 0.1060 \\
 & VILA-U & 61.92 & 44.77 & 0.3595 &  & VILA-U & 48.10 & 62.15 & 0.1469 \\\midrule
\multirow{6}{*}{SDXL} & Flux & 53.97 & 44.77 & 0.3191 & \multirow{6}{*}{Qwen2VL} & LLaVA & 48.51 & 64.63 & 0.1498 \\
 & \textcolor{paircolor}{SDXL} & \textcolor{paircolor}{53.97} & \textcolor{paircolor}{44.77} &  \textcolor{paircolor}{0.3538} &  & \textcolor{paircolor}{Qwen2VL} & \textcolor{paircolor}{48.51} & \textcolor{paircolor}{64.63} &  \textcolor{paircolor}{0.0682} \\
 & Chameleon & 53.97 & 44.77 & 0.3642 &  & Chameleon & 48.51 & 64.63 & 0.1552 \\
 & Emu3 & 53.97 & 44.77 & 0.3665 &  & Emu3 & 48.51 & 64.63 & 0.0882 \\
 & Seed-X & 53.97 & 44.77 & 0.3883 &  & Seed-X & 48.51 & 64.63 & 0.0259 \\
 & VILA-U & 53.97 & 44.77 & 0.3848 &  & VILA-U & 48.51 & 64.63 & 0.0585 \\\midrule
\multirow{6}{*}{Chameleon} & Flux & 51.24 & 42.26 & 0.3096 & \multirow{6}{*}{Chameleon} & LLaVA & 47.69 & 54.30 & 0.0323 \\
 & SDXL & 51.24 & 42.26 & 0.3096 &  & Qwen2VL & 47.69 & 54.30 & 0.1053 \\
 & \textcolor{paircolor}{Chameleon} & \textcolor{paircolor}{51.24} & \textcolor{paircolor}{42.26} &  \textcolor{paircolor}{0.3768} &  & \textcolor{paircolor}{Chameleon} & \textcolor{paircolor}{47.69} & \textcolor{paircolor}{54.30} &  \textcolor{paircolor}{0.0688} \\
 & Emu3 & 51.24 & 42.26 & 0.4132 &  & Emu3 & 47.69 & 54.30 & -0.0413 \\
 & Seed-X & 51.24 & 42.26 & 0.4188 &  & Seed-X & 47.69 & 54.30 & -0.0600 \\
 & VILA-U & 51.24 & 42.26 & 0.4538 &  & VILA-U & 47.69 & 54.30 & 0.1540 \\\midrule
\multirow{6}{*}{Emu3} & Flux & 56.07 & 49.37 & 0.3325 & \multirow{6}{*}{Emu3} & LLaVA & 56.82 & 53.47 & 0.4055 \\
 & SDXL & 56.07 & 49.37 & 0.2798 &  & Qwen2VL & 56.82 & 53.47 & 0.3658 \\
 & Chameleon & 56.07 & 49.37 & 0.3580 &  & Chameleon & 56.82 & 53.47 & 0.2568 \\
 & \textcolor{paircolor}{Emu3} & \textcolor{paircolor}{56.07} & \textcolor{paircolor}{49.37} &  \textcolor{paircolor}{0.4052} &  & \textcolor{paircolor}{Emu3} & \textcolor{paircolor}{56.82} & \textcolor{paircolor}{53.47} &  \textcolor{paircolor}{0.3015} \\
 & Seed-X & 56.07 & 49.37 & 0.3991 &  & Seed-X & 56.82 & 53.47 & 0.3911 \\
 & VILA-U & 56.07 & 49.37 & 0.3801 &  & VILA-U & 56.82 & 53.47 & 0.3604 \\\midrule
\multirow{6}{*}{Seed-X} & Flux & 57.74 & 46.86 & 0.3729 & \multirow{6}{*}{Seed-X} & LLaVA & 62.26 & 63.80 & 0.4019 \\
 & SDXL & 57.74 & 46.86 & 0.3905 &  & Qwen2VL & 62.26 & 63.80 & 0.3963 \\
 & Chameleon & 57.74 & 46.86 & 0.3103 &  & Chameleon & 62.26 & 63.80 & 0.2899 \\
 & Emu3 & 57.74 & 46.86 & 0.3496 &  & Emu3 & 62.26 & 63.80 & 0.3320 \\
 & \textcolor{paircolor}{Seed-X} & \textcolor{paircolor}{57.74} & \textcolor{paircolor}{46.86} &  \textcolor{paircolor}{0.3340} &  & \textcolor{paircolor}{Seed-X} & \textcolor{paircolor}{62.26} & \textcolor{paircolor}{63.80} &  \textcolor{paircolor}{0.4593} \\
 & VILA-U & 57.74 & 46.86 & 0.3502 &  & VILA-U & 62.26 & 63.80 & 0.3350 \\\midrule
\multirow{6}{*}{VILA-U} & Flux & 53.56 & 43.10 & 0.3947 & \multirow{6}{*}{VILA-U} & LLaVA & 58.91 & 55.12 & 0.3424 \\
 & SDXL & 53.56 & 43.10 & 0.2908 &  & Qwen2VL & 58.91 & 55.12 & 0.3665 \\
 & Chameleon & 53.56 & 43.10 & 0.3879 &  & Chameleon & 58.91 & 55.12 & 0.1777 \\
 & Emu3 & 53.56 & 43.10 & 0.4123 &  & Emu3 & 58.91 & 55.12 & 0.3129 \\
 & Seed-X & 53.56 & 43.10 & 0.4463 &  & Seed-X & 58.91 & 55.12 & 0.3271 \\
 & \textcolor{paircolor}{VILA-U} & \textcolor{paircolor}{53.56} & \textcolor{paircolor}{43.10} &  \textcolor{paircolor}{0.4085} &  & \textcolor{paircolor}{VILA-U} & \textcolor{paircolor}{58.91} & \textcolor{paircolor}{55.12} &  \textcolor{paircolor}{0.4340} \\\bottomrule

\end{tabular}}
\caption{Experimental results for testing forward equivariance.}
\label{tab:equiv}
\end{table*}

%% file: tables/sym.tex
\begin{table*}[h!]
\centering
\resizebox{0.98\textwidth}{!}{
\begin{tabular}{llcc|llcc}
\toprule
\multicolumn{4}{c|}{\textbf{Image}} & \multicolumn{4}{c}{\textbf{Text}} \\
\textbf{I2T} & \textbf{T2I} & \textbf{Accuracy (\%)} & \textbf{F1 (\%)} & \textbf{T2I} & \textbf{I2T} & \textbf{Accuracy (\%)} & \textbf{F1 (\%)} \\ \toprule
\multirow{6}{*}{LLaVA} & Flux & 45.61 & 51.85 & \multirow{6}{*}{Flux} & LLaVA & 57.85 & 62.22 \\
 & SDXL & 41.42 & 47.37 &  & Qwen2VL & 65.29 & 70.83 \\
 & Chameleon & 44.21 & 50.18 &  & Chameleon & 54.13 & 58.43 \\
 & Emu3 & 48.95 & 58.22 &  & Emu3 & 58.58 & 62.64 \\
 & Seed-X & 46.86 & 56.06 &  & Seed-X & 60.25 & 65.20 \\
 & VILA-U & 50.63 & 59.59 &  & VILA-U & 57.32 & 62.22 \\\midrule
\multirow{6}{*}{Qwen2VL} & Flux & 39.75 & 45.04 & \multirow{6}{*}{SDXL} & LLaVA & 57.02 & 60.61 \\
 & SDXL & 43.10 & 48.48 &  & Qwen2VL & 57.44 & 61.71 \\
 & Chameleon & 43.80 & 49.63 &  & Chameleon & 52.07 & 52.85 \\
 & Emu3 & 46.44 & 55.86 &  & Emu3 & 55.65 & 57.60 \\
 & Seed-X & 46.03 & 54.09 &  & Seed-X & 57.74 & 61.30 \\
 & VILA-U & 48.54 & 57.14 &  & VILA-U & 56.90 & 60.84 \\\midrule
\multirow{6}{*}{Chameleon} & Flux & 45.61 & 52.55 & \multirow{6}{*}{Chameleon} & LLaVA & 54.13 & 56.13 \\
 & SDXL & 46.86 & 52.43 &  & Qwen2VL & 55.79 & 57.03 \\
 & \textcolor{paircolor}{Chameleon} & \textcolor{paircolor}{41.32} & \textcolor{paircolor}{45.80} &  & \textcolor{paircolor}{Chameleon} & \textcolor{paircolor}{56.20} & \textcolor{paircolor}{58.91} \\
 & Emu3 & 48.54 & 56.54 &  & Emu3 & 55.23 & 56.68 \\
 & Seed-X & 45.19 & 52.01 &  & Seed-X & 55.65 & 57.94 \\
 & VILA-U & 46.03 & 52.75 &  & VILA-U & 54.81 & 57.48 \\\midrule
\multirow{6}{*}{Emu3} & Flux & 53.14 & 62.16 & \multirow{6}{*}{Emu3} & LLaVA & 56.20 & 59.85 \\
 & SDXL & 50.63 & 58.74 &  & Qwen2VL & 54.55 & 58.02 \\
 & Chameleon & 53.31 & 60.90 &  & Chameleon & 57.02 & 60.31 \\
 & \textcolor{paircolor}{Emu3} & \textcolor{paircolor}{62.76} & \textcolor{paircolor}{71.01} &  & \textcolor{paircolor}{Emu3} & \textcolor{paircolor}{56.90} & \textcolor{paircolor}{60.54} \\
 & Seed-X & 57.32 & 66.00 &  & Seed-X & 55.65 & 58.59 \\
 & VILA-U & 51.88 & 61.28 &  & VILA-U & 56.07 & 60.67 \\\midrule
\multirow{6}{*}{Seed-X} & Flux & 56.90 & 66.67 & \multirow{6}{*}{Seed-X} & LLaVA & 57.44 & 60.84 \\
 & SDXL & 56.07 & 64.65 &  & Qwen2VL & 54.55 & 58.96 \\
 & Chameleon & 52.07 & 59.72 &  & Chameleon & 56.61 & 59.14 \\
 & Emu3 & 55.23 & 65.37 &  & Emu3 & 54.81 & 56.45 \\
 & \textcolor{paircolor}{Seed-X} & \textcolor{paircolor}{59.83} & \textcolor{paircolor}{69.43} &  & \textcolor{paircolor}{Seed-X} & \textcolor{paircolor}{56.90} & \textcolor{paircolor}{61.13} \\
 & VILA-U & 60.25 & 70.40 &  & VILA-U & 52.72 & 56.70 \\\midrule
\multirow{6}{*}{VILA-U} & Flux & 58.58 & 68.17 & \multirow{6}{*}{VILA-U} & LLaVA & 55.79 & 59.32 \\
 & SDXL & 54.81 & 63.76 &  & Qwen2VL & 56.20 & 60.15 \\
 & Chameleon & 46.69 & 54.74 &  & Chameleon & 54.55 & 57.69 \\
 & Emu3 & 51.46 & 61.33 &  & Emu3 & 51.05 & 50.63 \\
 & Seed-X & 55.23 & 65.59 &  & Seed-X & 56.07 & 59.77 \\
 & \textcolor{paircolor}{VILA-U} & \textcolor{paircolor}{56.90} & \textcolor{paircolor}{67.09} &  & \textcolor{paircolor}{VILA-U} & \textcolor{paircolor}{56.07} & \textcolor{paircolor}{59.14} \\\bottomrule

\end{tabular}}
\caption{Experimental results for testing conjugated equivariance.}
\label{tab:sym}
\end{table*}

%% file: sections/ax_prompts.tex
\section{Instruction Prompts}
\label{sec:ax_prompts}

All instruction-following models (LLaVA-Next, Qwen2VL, Chameleon, Emu3, VILA-U, and SEED-X) used in our experiments come with predefined instruction templates, which we adhere to. For verbatim templates, refer to the respective repositories. Below, we detail the instruction prompts employed for various tasks in our experiments.

\paragraph{Models}
General-purpose instructions used in the experiments include:
\begin{itemize}
\item \textbf{Image-to-Text:} \texttt{Describe the image in detail.}
\item \textbf{Text-to-Image:} \texttt{Create an image from the following text.}
\end{itemize}

Certain models require task-specific prompts, as per their official implementations:
\begin{itemize}
    \item \textbf{Seed-X (Image-to-Text):} \texttt{Generate an image: {caption}}
\end{itemize}

For Emu3, the official implementation recommends using a set of positive and negative prompts for classifier-free guidance sampling in image generation:
\begin{itemize}
\item \textbf{Positive:} \texttt{{caption} masterpiece, film grained, best quality.}
\item \textbf{Negative:} \texttt{lowres, bad anatomy, bad hands, text, error, missing fingers, extra digit, fewer digits, cropped, worst quality, low quality, normal quality, jpeg artifacts, signature, watermark, username, blurry.}
\end{itemize}

\paragraph{Evaluation}
For evaluation, we directly use the questions as prompts without additional instructions. Empirically, this approach yields the highest correctness, as our evaluator, PaliGemma, is already trained for binary VQA tasks.

\section{Acknowledgements}
\label{sec:ack}

The entirety of this paper, including the main manuscript and supplementary materials, was prepared with the assistance of a GPT-based AI assistant. The use of the AI assistant was strictly limited to improving the language of the initial hand-written draft and automating table formatting. Additionally, the accompanying code for this paper was developed with the support of an AI assistant (CoPilot).

\paragraph{Figures}
All icons used in the figures are from \url{www.flaticon.com}. All images are from our \modelname dataset, except for~\cref{fig:intuition}.

%% file: main.bbl
\begin{thebibliography}{49}
\providecommand{\natexlab}[1]{#1}

\bibitem[{Ananthram et~al.(2024)Ananthram, Stengel-Eskin, Vondrick, Bansal, and McKeown}]{ananthram2024see}
Amith Ananthram, Elias Stengel-Eskin, Carl Vondrick, Mohit Bansal, and Kathleen McKeown. 2024.
\newblock See it from my perspective: Diagnosing the western cultural bias of large vision-language models in image understanding.
\newblock \emph{arXiv preprint arXiv:2406.11665}.

\bibitem[{Bai et~al.(2023)Bai, Bai, Yang, Wang, Tan, Wang, Lin, Zhou, and Zhou}]{bai2023qwen}
Jinze Bai, Shuai Bai, Shusheng Yang, Shijie Wang, Sinan Tan, Peng Wang, Junyang Lin, Chang Zhou, and Jingren Zhou. 2023.
\newblock Qwen-vl: A versatile vision-language model for understanding, localization, text reading, and beyond.
\newblock \emph{arXiv preprint arXiv:2308.12966}, 1(2):3.

\bibitem[{Bent(2024)}]{bent2024semantic}
Brinnae Bent. 2024.
\newblock Semantic approach to quantifying the consistency of diffusion model image generation.
\newblock \emph{arXiv preprint arXiv:2404.08799}.

\bibitem[{Betker et~al.(2023)Betker, Goh, Jing, Brooks, Wang, Li, Ouyang, Zhuang, Lee, Guo et~al.}]{betker2023improving}
James Betker, Gabriel Goh, Li~Jing, Tim Brooks, Jianfeng Wang, Linjie Li, Long Ouyang, Juntang Zhuang, Joyce Lee, Yufei Guo, et~al. 2023.
\newblock Improving image generation with better captions.
\newblock \emph{Computer Science. https://cdn. openai. com/papers/dall-e-3. pdf}, 2(3):8.

\bibitem[{Bielawski and VanRullen(2023)}]{bielawski2023clip}
Romain Bielawski and Rufin VanRullen. 2023.
\newblock Clip-based image captioning via unsupervised cycle-consistency in the latent space.
\newblock In \emph{8th Workshop on Representation Learning for NLP (RepL4NLP 2023)}, pages 266--275. Association for Computational Linguistics.

\bibitem[{Chen et~al.(2015)Chen, Fang, Lin, Vedantam, Gupta, Doll{\'a}r, and Zitnick}]{chen2015microsoft}
Xinlei Chen, Hao Fang, Tsung-Yi Lin, Ramakrishna Vedantam, Saurabh Gupta, Piotr Doll{\'a}r, and C~Lawrence Zitnick. 2015.
\newblock Microsoft coco captions: Data collection and evaluation server.
\newblock \emph{arXiv preprint arXiv:1504.00325}.

\bibitem[{Chern et~al.(2024)Chern, Su, Ma, and Liu}]{chern2024anole}
Ethan Chern, Jiadi Su, Yan Ma, and Pengfei Liu. 2024.
\newblock Anole: An open, autoregressive, native large multimodal models for interleaved image-text generation.
\newblock \emph{arXiv preprint arXiv:2407.06135}.

\bibitem[{Cho et~al.(2024)Cho, Hu, Baldridge, Garg, Anderson, Krishna, Bansal, Pont-Tuset, and Wang}]{chodavidsonian}
Jaemin Cho, Yushi Hu, Jason~Michael Baldridge, Roopal Garg, Peter Anderson, Ranjay Krishna, Mohit Bansal, Jordi Pont-Tuset, and Su~Wang. 2024.
\newblock Davidsonian scene graph: Improving reliability in fine-grained evaluation for text-to-image generation.
\newblock In \emph{The Twelfth International Conference on Learning Representations}.

\bibitem[{Chou et~al.(2024)Chou, Chandhok, Little, and Sigal}]{chou2024mm}
Shih-Han Chou, Shivam Chandhok, James~J Little, and Leonid Sigal. 2024.
\newblock $\text{MM-R}^3$: On (in-) consistency of multi-modal large language models (mllms).
\newblock \emph{arXiv preprint arXiv:2410.04778}.

\bibitem[{Cohen and Welling(2016)}]{cohen2016group}
Taco Cohen and Max Welling. 2016.
\newblock Group equivariant convolutional networks.
\newblock In \emph{International conference on machine learning}, pages 2990--2999. PMLR.

\bibitem[{Diesendruck et~al.(2024)Diesendruck, Lin, Imani, Mahalingam, Xu, and Zhao}]{diesendruck2024learning}
Maurice Diesendruck, Jianzhe Lin, Shima Imani, Gayathri Mahalingam, Mingyang Xu, and Jie Zhao. 2024.
\newblock Learning how to ask: Cycle-consistency refines prompts in multimodal foundation models.
\newblock \emph{arXiv preprint arXiv:2402.08756}.

\bibitem[{Ge et~al.(2024{\natexlab{a}})Ge, Zhao, Zeng, Ge, Li, Wang, and Shan}]{ge2024making}
Yuying Ge, Sijie Zhao, Ziyun Zeng, Yixiao Ge, Chen Li, Xintao Wang, and Ying Shan. 2024{\natexlab{a}}.
\newblock \href {https://openreview.net/forum?id=0Nui91LBQS} {Making {LL}a{MA} {SEE} and draw with {SEED} tokenizer}.
\newblock In \emph{The Twelfth International Conference on Learning Representations}.

\bibitem[{Ge et~al.(2024{\natexlab{b}})Ge, Zhao, Zhu, Ge, Yi, Song, Li, Ding, and Shan}]{ge2024seed}
Yuying Ge, Sijie Zhao, Jinguo Zhu, Yixiao Ge, Kun Yi, Lin Song, Chen Li, Xiaohan Ding, and Ying Shan. 2024{\natexlab{b}}.
\newblock Seed-x: Multimodal models with unified multi-granularity comprehension and generation.
\newblock \emph{arXiv preprint arXiv:2404.14396}.

\bibitem[{{Google DeepMind}(2023)}]{imagen2}
{Google DeepMind}. 2023.
\newblock \href {https://deepmind.google/technologies/imagen-2/} {Imagen 2}.

\bibitem[{Hessel et~al.(2021)Hessel, Holtzman, Forbes, Bras, and Choi}]{hessel2021clipscore}
Jack Hessel, Ari Holtzman, Maxwell Forbes, Ronan~Le Bras, and Yejin Choi. 2021.
\newblock Clipscore: A reference-free evaluation metric for image captioning.
\newblock \emph{arXiv preprint arXiv:2104.08718}.

\bibitem[{Hsieh et~al.(2024)Hsieh, Hsieh, Yeh, B{\'e}thune, Ansari, Vasu, Li, Krishna, Tuzel, and Cuturi}]{hsieh2024graph}
Yu-Guan Hsieh, Cheng-Yu Hsieh, Shih-Ying Yeh, Louis B{\'e}thune, Hadi~Pour Ansari, Pavan Kumar~Anasosalu Vasu, Chun-Liang Li, Ranjay Krishna, Oncel Tuzel, and Marco Cuturi. 2024.
\newblock Graph-based captioning: Enhancing visual descriptions by interconnecting region captions.
\newblock \emph{arXiv preprint arXiv:2407.06723}.

\bibitem[{Hu et~al.(2023)Hu, Liu, Kasai, Wang, Ostendorf, Krishna, and Smith}]{hu2023tifa}
Yushi Hu, Benlin Liu, Jungo Kasai, Yizhong Wang, Mari Ostendorf, Ranjay Krishna, and Noah~A Smith. 2023.
\newblock Tifa: Accurate and interpretable text-to-image faithfulness evaluation with question answering.
\newblock In \emph{Proceedings of the IEEE/CVF International Conference on Computer Vision}, pages 20406--20417.

\bibitem[{Huang et~al.(2021)Huang, Du, Xue, Chen, Zhao, and Huang}]{huang2021makes}
Yu~Huang, Chenzhuang Du, Zihui Xue, Xuanyao Chen, Hang Zhao, and Longbo Huang. 2021.
\newblock What makes multi-modal learning better than single (provably).
\newblock \emph{Advances in Neural Information Processing Systems}, 34:10944--10956.

\bibitem[{Jin et~al.(2023)Jin, Xu, Chen, Liao, Tan, Chen, Lei, Liu, Song, Lei et~al.}]{jin2023unified}
Yang Jin, Kun Xu, Liwei Chen, Chao Liao, Jianchao Tan, Bin Chen, Chenyi Lei, An~Liu, Chengru Song, Xiaoqiang Lei, et~al. 2023.
\newblock Unified language-vision pretraining with dynamic discrete visual tokenization.
\newblock \emph{arXiv preprint arXiv:2309.04669}.

\bibitem[{Kim et~al.(2019)Kim, Kitaev, Chen, Rohrbach, Zhang, Tian, Batra, and Parikh}]{kim2019codraw}
Jin-Hwa Kim, Nikita Kitaev, Xinlei Chen, Marcus Rohrbach, Byoung-Tak Zhang, Yuandong Tian, Dhruv Batra, and Devi Parikh. 2019.
\newblock Codraw: Collaborative drawing as a testbed for grounded goal-driven communication.
\newblock In \emph{Proceedings of the 57th Annual Meeting of the Association for Computational Linguistics}, pages 6495--6513.

\bibitem[{Labs(2024)}]{blackforest2024flux}
Black~Forest Labs. 2024.
\newblock Flux.1: Advanced text-to-image generation.
\newblock \url{https://blackforestlabs.ai/flux-1/}.

\bibitem[{Lin et~al.(2024)Lin, Pathak, Li, Li, Xia, Neubig, Zhang, and Ramanan}]{lin2024evaluating}
Zhiqiu Lin, Deepak Pathak, Baiqi Li, Jiayao Li, Xide Xia, Graham Neubig, Pengchuan Zhang, and Deva Ramanan. 2024.
\newblock Evaluating text-to-visual generation with image-to-text generation.
\newblock In \emph{European Conference on Computer Vision}, pages 366--384. Springer.

\bibitem[{Liu et~al.(2024)Liu, Li, Li, and Lee}]{liu2024improved}
Haotian Liu, Chunyuan Li, Yuheng Li, and Yong~Jae Lee. 2024.
\newblock Improved baselines with visual instruction tuning.
\newblock In \emph{Proceedings of the IEEE/CVF Conference on Computer Vision and Pattern Recognition}, pages 26296--26306.

\bibitem[{Lu et~al.(2024)Lu, Clark, Lee, Zhang, Khosla, Marten, Hoiem, and Kembhavi}]{lu2024unified}
Jiasen Lu, Christopher Clark, Sangho Lee, Zichen Zhang, Savya Khosla, Ryan Marten, Derek Hoiem, and Aniruddha Kembhavi. 2024.
\newblock Unified-io 2: Scaling autoregressive multimodal models with vision language audio and action.
\newblock In \emph{Proceedings of the IEEE/CVF Conference on Computer Vision and Pattern Recognition}, pages 26439--26455.

\bibitem[{Lu(2024)}]{erwold-2024-qwen2vl-flux}
Pengqi Lu. 2024.
\newblock \href {https://github.com/erwold/qwen2vl-flux} {Qwen2vl-flux: Unifying image and text guidance for controllable image generation}.

\bibitem[{Lu(2023)}]{lu2023theory}
Zhou Lu. 2023.
\newblock A theory of multimodal learning.
\newblock \emph{Advances in Neural Information Processing Systems}, 36:57244--57255.

\bibitem[{Ma et~al.(2023)Ma, Hong, Gul, Gandhi, Gao, and Krishna}]{ma2023crepe}
Zixian Ma, Jerry Hong, Mustafa~Omer Gul, Mona Gandhi, Irena Gao, and Ranjay Krishna. 2023.
\newblock Crepe: Can vision-language foundation models reason compositionally?
\newblock In \emph{Proceedings of the IEEE/CVF Conference on Computer Vision and Pattern Recognition}, pages 10910--10921.

\bibitem[{Nisbett et~al.(2001)Nisbett, Peng, Choi, and Norenzayan}]{nisbett2001culture}
Richard~E Nisbett, Kaiping Peng, Incheol Choi, and Ara Norenzayan. 2001.
\newblock Culture and systems of thought: holistic versus analytic cognition.
\newblock \emph{Psychological review}, 108(2):291.

\bibitem[{OpenAI(2023)}]{openai2023gpt4}
OpenAI. 2023.
\newblock Gpt-4: Openai’s large multimodal model.
\newblock \url{https://openai.com/research/gpt-4}.

\bibitem[{Pan et~al.(2024)Pan, Dong, Huang, Peng, Chen, and Wei}]{pan2024kosmosg}
Xichen Pan, Li~Dong, Shaohan Huang, Zhiliang Peng, Wenhu Chen, and Furu Wei. 2024.
\newblock \href {https://openreview.net/forum?id=he6mX9LTyE} {Kosmos-g: Generating images in context with multimodal large language models}.
\newblock In \emph{The Twelfth International Conference on Learning Representations}.

\bibitem[{Podell et~al.(2024)Podell, English, Lacey, Blattmann, Dockhorn, M{\"u}ller, Penna, and Rombach}]{podellsdxl}
Dustin Podell, Zion English, Kyle Lacey, Andreas Blattmann, Tim Dockhorn, Jonas M{\"u}ller, Joe Penna, and Robin Rombach. 2024.
\newblock Sdxl: Improving latent diffusion models for high-resolution image synthesis.
\newblock In \emph{The Twelfth International Conference on Learning Representations}.

\bibitem[{Radford et~al.(2021)Radford, Kim, Hallacy, Ramesh, Goh, Agarwal, Sastry, Askell, Mishkin, Clark et~al.}]{radford2021learning}
Alec Radford, Jong~Wook Kim, Chris Hallacy, Aditya Ramesh, Gabriel Goh, Sandhini Agarwal, Girish Sastry, Amanda Askell, Pamela Mishkin, Jack Clark, et~al. 2021.
\newblock Learning transferable visual models from natural language supervision.
\newblock In \emph{International conference on machine learning}, pages 8748--8763. PMLR.

\bibitem[{Rombach et~al.(2022)Rombach, Blattmann, Lorenz, Esser, and Ommer}]{rombach2022stable}
Robin Rombach, Andreas Blattmann, Dominik Lorenz, Patrick Esser, and Bj{\"o}rn Ommer. 2022.
\newblock \href {https://arxiv.org/abs/2112.10752} {High-resolution image synthesis with latent diffusion models}.
\newblock In \emph{Proceedings of the IEEE/CVF Conference on Computer Vision and Pattern Recognition (CVPR)}, pages 10684--10695.

\bibitem[{Steiner et~al.(2024)Steiner, Pinto, Tschannen, Keysers, Wang, Bitton, Gritsenko, Minderer, Sherbondy, Long et~al.}]{steiner2024paligemma}
Andreas Steiner, Andr{\'e}~Susano Pinto, Michael Tschannen, Daniel Keysers, Xiao Wang, Yonatan Bitton, Alexey Gritsenko, Matthias Minderer, Anthony Sherbondy, Shangbang Long, et~al. 2024.
\newblock Paligemma 2: A family of versatile vlms for transfer.
\newblock \emph{arXiv preprint arXiv:2412.03555}.

\bibitem[{Sun et~al.(2024{\natexlab{a}})Sun, Cui, Zhang, Zhang, Yu, Wang, Rao, Liu, Huang, and Wang}]{sun2024generative}
Quan Sun, Yufeng Cui, Xiaosong Zhang, Fan Zhang, Qiying Yu, Yueze Wang, Yongming Rao, Jingjing Liu, Tiejun Huang, and Xinlong Wang. 2024{\natexlab{a}}.
\newblock Generative multimodal models are in-context learners.
\newblock In \emph{Proceedings of the IEEE/CVF Conference on Computer Vision and Pattern Recognition}, pages 14398--14409.

\bibitem[{Sun et~al.(2024{\natexlab{b}})Sun, Chu, Qin, and Ren}]{sun2024prompt}
Yichen Sun, Zhixuan Chu, Zhan Qin, and Kui Ren. 2024{\natexlab{b}}.
\newblock Prompt-consistency image generation (pcig): A unified framework integrating llms, knowledge graphs, and controllable diffusion models.
\newblock \emph{arXiv preprint arXiv:2406.16333}.

\bibitem[{Tan et~al.(2022)Tan, Yang, Ye, Wang, Yan, Nguyen, and Huang}]{tan2022ssd}
Zhaorui Tan, Xi~Yang, Zihan Ye, Qiufeng Wang, Yuyao Yan, Anh Nguyen, and Kaizhu Huang. 2022.
\newblock Ssd: Towards better text-image consistency metric in text-to-image generation.
\newblock \emph{arXiv preprint arXiv:2210.15235}.

\bibitem[{Team(2024)}]{team2024chameleon}
Chameleon Team. 2024.
\newblock Chameleon: Mixed-modal early-fusion foundation models.
\newblock \emph{arXiv preprint arXiv:2405.09818}.

\bibitem[{Tewel et~al.(2024)Tewel, Kaduri, Gal, Kasten, Wolf, Chechik, and Atzmon}]{tewel2024training}
Yoad Tewel, Omri Kaduri, Rinon Gal, Yoni Kasten, Lior Wolf, Gal Chechik, and Yuval Atzmon. 2024.
\newblock Training-free consistent text-to-image generation.
\newblock \emph{ACM Transactions on Graphics (TOG)}, 43(4):1--18.

\bibitem[{Tong et~al.(2024)Tong, Liu, Zhai, Ma, LeCun, and Xie}]{tong2024eyes}
Shengbang Tong, Zhuang Liu, Yuexiang Zhai, Yi~Ma, Yann LeCun, and Saining Xie. 2024.
\newblock Eyes wide shut? exploring the visual shortcomings of multimodal llms.
\newblock In \emph{Proceedings of the IEEE/CVF Conference on Computer Vision and Pattern Recognition}, pages 9568--9578.

\bibitem[{van~den Oord et~al.(2017)van~den Oord, Vinyals, and Kavukcuoglu}]{oord2017vqvae}
Aaron van~den Oord, Oriol Vinyals, and Koray Kavukcuoglu. 2017.
\newblock Neural discrete representation learning.
\newblock In \emph{Advances in Neural Information Processing Systems (NeurIPS)}, volume~30.

\bibitem[{Wang et~al.(2022)Wang, Yang, Men, Lin, Bai, Li, Ma, Zhou, Zhou, and Yang}]{wang2022ofa}
Peng Wang, An~Yang, Rui Men, Junyang Lin, Shuai Bai, Zhikang Li, Jianxin Ma, Chang Zhou, Jingren Zhou, and Hongxia Yang. 2022.
\newblock Ofa: Unifying architectures, tasks, and modalities through a simple sequence-to-sequence learning framework.
\newblock In \emph{International conference on machine learning}, pages 23318--23340. PMLR.

\bibitem[{Wang et~al.(2024)Wang, Zhang, Luo, Sun, Cui, Wang, Zhang, Wang, Li, Yu et~al.}]{wang2024emu3}
Xinlong Wang, Xiaosong Zhang, Zhengxiong Luo, Quan Sun, Yufeng Cui, Jinsheng Wang, Fan Zhang, Yueze Wang, Zhen Li, Qiying Yu, et~al. 2024.
\newblock Emu3: Next-token prediction is all you need.
\newblock \emph{arXiv preprint arXiv:2409.18869}.

\bibitem[{Wu et~al.(2024)Wu, Zhang, Chen, Tang, Li, Fang, Zhu, Xie, Yin, Yi et~al.}]{wu2024vila}
Yecheng Wu, Zhuoyang Zhang, Junyu Chen, Haotian Tang, Dacheng Li, Yunhao Fang, Ligeng Zhu, Enze Xie, Hongxu Yin, Li~Yi, et~al. 2024.
\newblock Vila-u: a unified foundation model integrating visual understanding and generation.
\newblock \emph{arXiv preprint arXiv:2409.04429}.

\bibitem[{Yang et~al.(2024)Yang, Yang, Hui, Zheng, Yu, Zhou, Li, Li, Liu, Huang et~al.}]{yang2024qwen2}
An~Yang, Baosong Yang, Binyuan Hui, Bo~Zheng, Bowen Yu, Chang Zhou, Chengpeng Li, Chengyuan Li, Dayiheng Liu, Fei Huang, et~al. 2024.
\newblock Qwen2 technical report.
\newblock \emph{arXiv preprint arXiv:2407.10671}.

\bibitem[{Zhang et~al.(2024{\natexlab{a}})Zhang, Pang, Du, Ren, Li, and Lin}]{zhang2024benchmarking}
Jiawei Zhang, Tianyu Pang, Chao Du, Yi~Ren, Bo~Li, and Min Lin. 2024{\natexlab{a}}.
\newblock Benchmarking large multimodal models against common corruptions.
\newblock \emph{arXiv preprint arXiv:2401.11943}.

\bibitem[{Zhang et~al.(2024{\natexlab{b}})Zhang, Zhang, Yin, Huang, Zhang, Hu, and Wan}]{zhang2024mc}
Junzhe Zhang, Huixuan Zhang, Xunjian Yin, Baizhou Huang, Xu~Zhang, Xinyu Hu, and Xiaojun Wan. 2024{\natexlab{b}}.
\newblock Mc-mke: A fine-grained multimodal knowledge editing benchmark emphasizing modality consistency.
\newblock \emph{arXiv preprint arXiv:2406.13219}.

\bibitem[{Zhou et~al.(2024)Zhou, Yu, Babu, Tirumala, Yasunaga, Shamis, Kahn, Ma, Zettlemoyer, and Levy}]{zhou2024transfusion}
Chunting Zhou, Lili Yu, Arun Babu, Kushal Tirumala, Michihiro Yasunaga, Leonid Shamis, Jacob Kahn, Xuezhe Ma, Luke Zettlemoyer, and Omer Levy. 2024.
\newblock Transfusion: Predict the next token and diffuse images with one multi-modal model.
\newblock \emph{arXiv preprint arXiv:2408.11039}.

\bibitem[{Zhu et~al.(2017)Zhu, Park, Isola, and Efros}]{zhu2017unpaired}
Jun-Yan Zhu, Taesung Park, Phillip Isola, and Alexei~A Efros. 2017.
\newblock Unpaired image-to-image translation using cycle-consistent adversarial networks.
\newblock In \emph{Proceedings of the IEEE international conference on computer vision}, pages 2223--2232.

\end{thebibliography}
